\documentclass[conference]{IEEEtran}
\IEEEoverridecommandlockouts

\usepackage{cite}
\usepackage{amsmath,amssymb,amsfonts}
\usepackage{algorithmic}
\usepackage{graphicx}
\usepackage{textcomp}
\usepackage{xcolor}
\usepackage{tikz}
\usepackage{bbm}
\usepackage{dsfont}
\usepackage{hyperref}

\usetikzlibrary{positioning, arrows.meta}
\usetikzlibrary{shapes.geometric, arrows.meta, positioning}

\def\BibTeX{{\rm B\kern-.05em{\sc i\kern-.025em b}\kern-.08em
    T\kern-.1667em\lower.7ex\hbox{E}\kern-.125emX}}

\begin{document}

\title{DinoLizer: Separating VAE and Diffusion Artifacts in Generative Inpainting Localization\\
}

\author{\IEEEauthorblockN{DOI Minh Thong, Vincent ITIER}
\IEEEauthorblockA{\textit{UMR 9189 CRIStAL, CERI SN} \\
\textit{IMT Nord Europe}\\
F-59000 Lille, France \\
minh-thong.doi@imt-nord-europe.fr \\
vincent.itier@imt-nord-europe.fr
}
\and
\IEEEauthorblockN{Jan BUTORA, Jérémie BOULANGER, Patrick BAS}
\IEEEauthorblockA{\textit{Univ. Lille, CNRS, Centrale Lille}\\
\textit{UMR 9189 CRIStAL} \\
F-59000 Lille, France \\
jeremie.boulanger@univ-lille.fr \\
\{jan.butora,patrick.bas\}@cnrs.fr
}


}

\maketitle

\begin{abstract}
We introduce DinoLizer, a DINOv2-based localizer of manipulated areas in generative inpainting.
 The model is trained to focus on semantically altered regions by treating reconstructed areas outside the inpainted mask as a separate class, which yields significant improvements w.r.t. the conventional approach. We train the model with LORA on the Query and Value of the transformer blocks and simply add 1 linear layer on top of the backbone to predict manipulations on a $14 \times 14$ patch resolution. Because DINOv2 only accepts fixed-sized images, we use a sliding window approach to aggregate the predictions on larger images. Empirical results show that DinoLizer outperforms state-of-the-art methods on our proposed dataset and SOTA inpainting datasets. Furthermore, it is very robust to JPEG (double) compression. On average, DinoLizer achieves a 20\% higher Intersection over Union score compared to the second best model. The code is publicly available here: \href{https://github.com/anonyme610/dinolizer}{https://github.com/anonyme610/dinolizer}.

\end{abstract}

\section{Introduction}

Generative Artificial Intelligence (Gen-AI) poses a significant threat to democracy by facilitating the creation of highly realistic disinformation. In particular, generative inpainting allows for the easy local removal or insertion of subjects by seamlessly blending authentic image components with synthetic backgrounds or objects. These hybrid manipulations are notoriously difficult to detect because the high fidelity of modern generative models makes it nearly impossible for human observers to differentiate between authentic and fabricated content.

Many existing inpainting methods are based either on diffusion models such as Stable-Diffusion~\cite{rombach2022high}, Glide~\cite{nichol2021glide}, or the commercial model from Adobe FireFly, but also other methods such as Fourier convolution-based methods and residual networks like LaMa~\cite{suvorov2022resolution} for object removal.
On diffusion models, the inpainting methods such as BrushNet~\cite{ju2024brushnet}, PowerPaint~\cite{zhuang2024task}, or HD-painter~\cite{manukyan2023hd} come from models that are fine-tuned or adapted using extra input channels. They are used either to infer a masked part of the image during the backward diffusion process (for object insertion) or to replace an object with another. To obtain a better inpainting quality, the alignment between masked regions and semantic objects can be enforced by using BLIP embeddings~\cite{xie2023smartbrush}. InpaintAnything~\cite{zhang2023adding} also proposes to mix the inpainting process with segmentation tools to perform high-quality inpainting.

Image forgery localization (IFL) methods have received increased attention due to rapidly evolving sophisticated algorithms for local image tampering.
 TruFor~\cite{turfor23} uses a Transformer-based fusion of high-level image information and a noise signature extracted by the Noiseprint++ feature extractor to predict local discrepancies. SAFIRE~\cite{safire25} segments an image into regions having the same properties, and localizes manipulations as outlier regions. ManTra-Net~\cite{mantra19} uses a fully convolutional network without any subsampling to predict local anomalies, and CAT-Net v2~\cite{catnetv2} is another CNN that focuses on distortion in JPEG compression artifacts due to forgeries. In~\cite{su25}, the authors aim to find non-semantic features useful for IFL, and modify SparseViT~\cite{sparse23} to extract context-irrelevant features by splitting the image into lattices of non-neighboring patches. However, most current methods have only been evaluated on traditional forgeries, such as copy-move and splicing, and not many works deal with AI-based inpainting with, e.g., diffusion models.

Our methodology for training DinoLizer is characterized by two core contributions. First, we prioritize computational efficiency by employing Low-Rank Adaptation (LoRA) \cite{hu2021loralowrankadaptationlarge} to fine-tune the DINOv2~\cite{dinov2} backbone, which significantly reduces the number of trainable parameters from 86M to 2.4M while preserving high performance. Second, we distinguish our approach by introducing a dedicated class for VAE artifacts; this enables the model to learn a more nuanced representation of generative traces rather than simply grouping them with the ``real'' class, thereby improving the distinction between authentic and manipulated regions.

\section{Related Work}
\subsection{Localization Methods Using Transformers}

Vision Transformers (ViTs) utilize self-attention mechanisms to capture long-range dependencies, making them highly effective for tasks such as object segmentation. Following the approach proposed by Segmenter~\cite{strudel21}, a common strategy involves an encoder-decoder architecture where the transformer serves as the encoder, and the decoder consists of a linear mapping between the predicted masks and patch embeddings, followed by upsampling. This methodology was similarly adopted by the DINOv2 ViT~\cite{dinov2} for image segmentation.

Regarding image forgery localization, SparseViT~\cite{su25} and DINOv3-IML~\cite{dinov3} are, to the best of our knowledge, the only works that utilize transformers for this purpose. SparseViT~\cite{su25} is trained on disjoint subsampled subsets of patches to enforce sparse self-attention maps; this prevents the model from focusing solely on semantic content, as many generation traces (e.g., VAE artifacts) are independent of image semantics. 
\begin{figure}

    \begin{center}

\begin{tikzpicture}[scale=0.65, transform shape,
    box/.style={rectangle, draw, thick, minimum height=1.5cm, minimum width=3cm, align=center, rounded corners},
    container/.style={rectangle, draw, ultra thick, dashed, inner sep=10pt, rounded corners},
    label_style/.style={font=\bfseries\small}
]

\node[container, color=red, draw=red, label_style, minimum width=9cm, minimum height=2.5cm] (binary_box) at (0,3.23) {};
\node[label_style, anchor=south] at (0, 3.85) {Binary Setup (Standard)};
\node[box, draw=red, fill=red!5] (bin_1) at (-2.5, 3) {\textcolor{red}{\bf VAE} / REAL};
\node[box, draw=red, fill=red!5] (bin_2) at (2.5, 3) {\textcolor{red}{\bf VAE} +  INPAINTING};

\node[container, color=teal, draw=teal, minimum width=11cm, minimum height=2.5cm] (multi_box) at (0,0.35) {};
\node[label_style, anchor=south] at (0, 1.0) {Multiclass Setup (Proposed)};
\node[box, draw=teal, fill=teal!5, minimum width=2cm] (multi_1) at (-3.5, 0.2) {REAL};
\node[box, draw=teal, fill=teal!5, minimum width=2cm] (multi_2) at (0, 0.2) {\textcolor{teal}{{\bf VAE}}};
\node[box, draw=teal, fill=teal!5, minimum width=3cm] (multi_3) at (3.5, 0.2) {\textcolor{teal}{\bf VAE} + INPAINTING};

\end{tikzpicture}

    \end{center}

\caption{
    In a binary setup, the learning mechanism can be confused by the VAE fingerprint, which is potentially present in both classes. On the contrary, in a multiclass setup, the VAE fingerprint is learned from one specific class, which eases the distinction with respect to inpainting.}
\label{fig:2and3classes}
\end{figure}

\subsection{Semantic and VAE bias removal}

Recently, different contributions have shown both excellent in-distribution and out-of-distribution detection ({\it i.e.} testing on unseen generators) performance by adopting training strategies designed to remove the semantic and/or VAE biases (artifacts introduced by the VAE encoder and decoder of generative models). The semantic bias usually comes from the fact that images are generated with prompts that are too short to represent a realistic description of a natural image. Another bias is due to the heterogeneity of VAEs that are used to decode the generated image from the latent to the pixel space. The training proposed in B-Free~\cite{bfree25} used only natural images going to the VAE or inpainted images with StableDiffusion2 to produce a training database of AI-processed images treated as ``generated" images. The original images come from the COCO~\cite{coco} database. For video detection, the frames of the generated class are also encoded by a VAE, and to improve generalization, the high-pass subbands of the original frames are injected in the encoded frames~\cite{corvi2025seeing}. A similar strategy~\cite{chen2025dual} has recently been proposed by using JPEG compression as a calibration between original images and auto-encoded images, the calibrated auto-encoded images forming the class of generated images during training. Note that all three methods~\cite{corvi2025seeing},\cite{bfree25} and~\cite{chen2025dual} finetune DINOv2 to train the classifier. 

\section{METHODOLOGY}

\subsection{Benchmark construction}

As the generative inpainting landscape evolves rapidly, existing datasets such as B-Free~\cite{bfree25} are becoming less representative of modern synthetic techniques. To provide a robust foundation for both training and evaluation, we require a dataset that captures the diversity of contemporary models while specifically challenging the ability of detectors to identify modifications across various scales.

\begin{figure}[!t]
    \centering
    \begin{tabular}{cc}
        \includegraphics[width=0.225\textwidth, trim=10pt 10pt 10pt 10pt,clip]{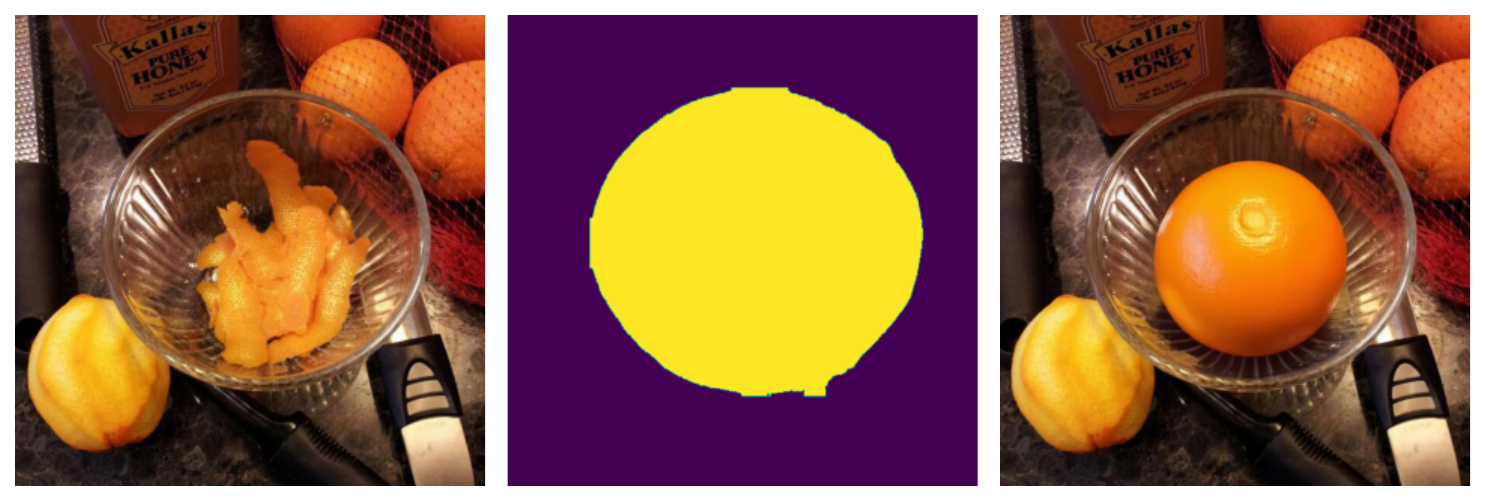} &
        \includegraphics[width=0.225\textwidth, trim=10pt 10pt 10pt 10pt,clip]{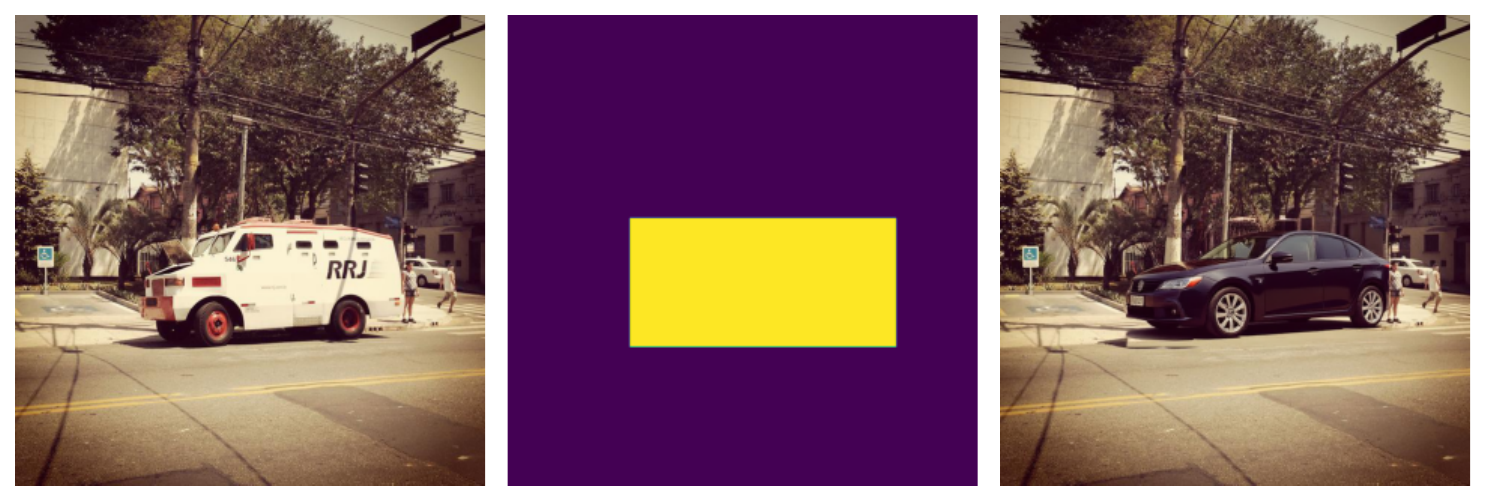} \\
        \includegraphics[width=0.225\textwidth, trim=10pt 10pt 10pt 10pt,clip]{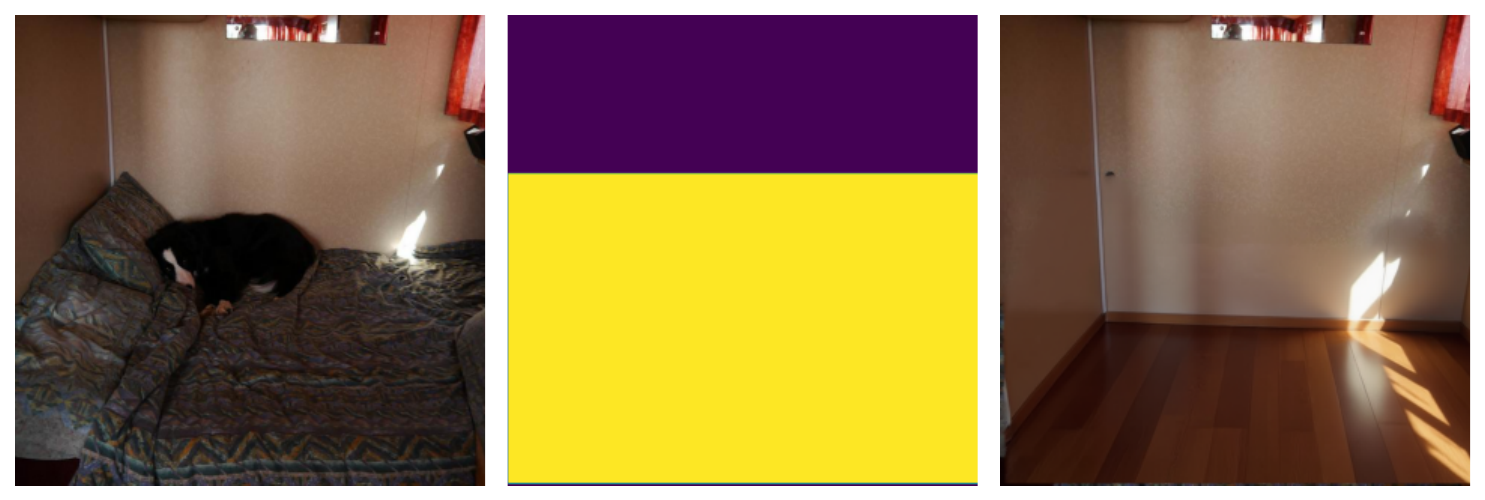} &
        \includegraphics[width=0.225\textwidth, trim=10pt 10pt 10pt 10pt,clip]{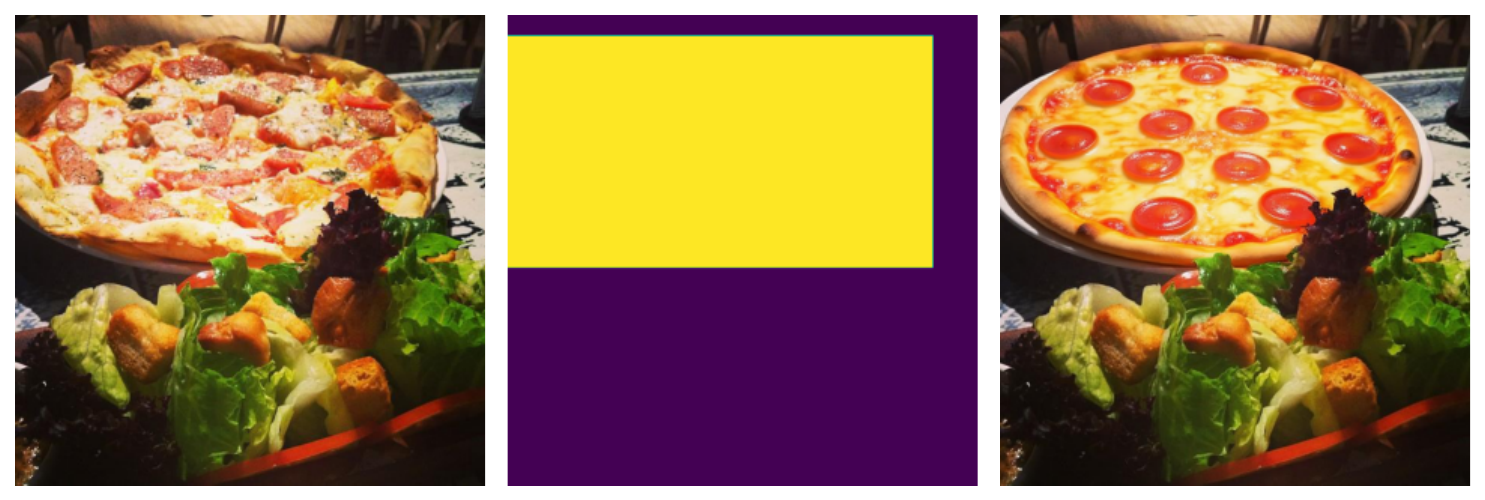} \\
     \end{tabular}
     \caption{Qualitative examples of the proposed dataset. For each example, we show the source image, the generated mask used for manipulation, and the final inpainted result (from left to right).}
    \label{fig:examples_grid}
\end{figure}

To increase diversity in source images and generative models, we introduce \textbf{FOSSIL (Foundation dataset for Occluded Structure Synthesis and Inpainting Localization)}, which pairs real images from the COCO dataset with tampered images generated by various SOTA inpainting models (e.g., DreamShaper8, SD1.5, SD2, SDXL, SD3, Flux Kontex, and Qwen Edit Image). The data construction pipeline involves two primary stages: (1) \textbf{Segmentation and Masking}, where we employ YOLOv11 to identify objects and select a mask whose area ratio $r$ is closest to a uniform sample $r \sim U(0, 1)$ to ensure variety in mask sizes; and (2) \textbf{Inpainting and Synthesis}, where objects are either removed or replaced with semantically similar counterparts. To maintain high-quality background consistency, 50\% of the images retain their original pristine backgrounds. The resulting dataset comprises 12,704 images ($\sim$2,000 per model), with qualitative examples shown in Figure~\ref{fig:examples_grid}. Note that while the proposed dataset is based on COCO, the methodology can be applied to any dataset at hand.
The dataset is publicly available here: \href{https://nextcloud.univ-lille.fr/index.php/s/FHTcLLwGggWHX6d}{FOSSIL}.

\subsection{Multiclass classification}

Most contemporary inpainting models employ a three-stage pipeline: first, the input image is encoded into a latent representation via a VAE encoder; second, the inpainting operation is performed within this latent space; and finally, the modified representation is decoded back into the pixel space using a VAE decoder. Consequently, when an image is generated by these models, the inpainted regions exhibit a superposition of both generative synthesis artifacts and reconstruction-based artifacts (VAE artifacts). In contrast, the regions outside the inpainted mask -- which undergo only reconstruction -- retain only VAE-specific artifacts.

Because the reconstructed regions maintain semantic content similar to the original image, they are typically treated as ``real'' in standard evaluation protocols. This leads to a conventional training setup involving only two classes: inpainted vs. not inpainted (semantically real). However, this configuration is suboptimal; by grouping VAE reconstructions with ``Real'' regions, we introduce conflicting supervisory signals. The model is forced to treat VAE-specific signatures as ``authentic'' in one class while simultaneously encountering those same signatures within the ``Inpainted'' class. To resolve this conflict, we adopt a three-class training strategy: Real, VAE, and Inpainted, as shown in Figure~\ref{fig:2and3classes}. This approach compels the model to learn a more granular and discriminative feature representation.

\subsection{Training Strategy}
We evaluate two distinct training paradigms in this work. The first employs the pre-trained DINOv2-B \cite{bfree25} model optimized for deepfake detection as a frozen feature extractor. In this configuration, we append either a simple linear layer (769 trainable parameters) or a sequence of two transformer blocks (14.2M trainable parameters) to the backbone.

The second approach leverages LoRA to achieve parameter-efficient fine-tuning. Specifically, we apply LoRA to the Query ($Q$) and Value ($V$) projections in each block of the DINOv2 architecture with a rank of $r=64$ and an alpha of $\alpha=64$, followed by a linear layer for patch-level classification. By keeping the majority of the model weights frozen, this method reduces the total number of trainable parameters to 2.4M.

Our primary focus is the second approach, which yields significantly better performance than the two-block configuration while remaining more parameter-efficient. Furthermore, we consider two types of backbones: \textbf{DINOv2} pretrained on LVD-142M \cite{dinov2} using self-supervised learning, and \textbf{DINOv2-B} \cite{bfree25} fine-tuned on the B-Free dataset for deepfake detection. We demonstrate that the DINOv2-B \cite{bfree25} backbone yields improvements in localization.

\subsection{ViT for Localization}
\label{sec:3class}
Since the \texttt{[CLS]} token aggregates information from the entire image, it is unsuitable for pixel-level localization. Instead, we focus on the patch embeddings (patch tokens) $[z_1^{(L)}; \ldots; z_N^{(L)}]$, each of which encodes the visual content of its corresponding patch. To obtain a per-patch modification probability $p_i$, we attach a lightweight linear classification head to each embedding and apply the softmax:
\begin{equation}
  p_i = \text{\it{softmax}}(W z_i^{(L)} + b), \quad i=1,\ldots, N, \label{eq:patch_prob}
\end{equation}
where $W z_i^{(L)} + b$ is the $i$-th patch logit.

Beyond the standard binary classification (real vs. inpainted), we introduce a third category: the \textbf{VAE class}. This class represents regions that were processed by the generative model but lie outside the primary mask, meaning they undergo regeneration without altering the underlying content. By explicitly classifying these generative traces as a separate category, we encourage the model to distinguish between general generative artifacts and the specific forged content of the inpainted object (see also Figure~\ref{fig:2and3classes}).

The model is trained with Weighted Cross Entropy loss,
with the class weights computed as:
  $w_0 = 1$, 
  $w_1 = \min\left(N_0/N_1, \tau \right)$, $w_2 = \min\left(N_0/N_2, \tau\right)$,
where $N_0$, $N_1$, and $N_2$ denote the number of pixels for class 0 (real), class 1 (inpainted), and class 2 (VAE), respectively. The stability constant $\tau$ is set to $50$ in our experiments.

\begin{table}[t]
\caption{Overview of evaluation datasets with their real image sources, inpainting generators (SD = Stable Diffusion, BtB: Beyond the Brush, COCOG: CocoGlide, MB = Magic Brush, MD = My Dataset). The presence of an original background (Orig Bg): \checkmark or a background that is auto-encoded: $\times$ in the forged images is also indicated.}
\label{tab:datasets_overview}
\centering
\resizebox{\columnwidth}{!}{
\begin{tabular}{|l|l|l|c|c|}
\hline
\textbf{Dataset} & \textbf{Real} & \textbf{Inpainting} & Orig & Fake\\
\ & \textbf{Source} & \textbf{Schemes} & Bg & Images\\
\hline
BtB~\cite{btb24} & Flickr30k & Fooocus~\cite{fooocus23}, SDXL & \checkmark & 7.4k\\

COCOG~\cite{turfor23} & MS-COCO~\cite{coco} & GLIDE~\cite{glide22} & \checkmark & 512\\

TGIF~\cite{tgif24} & MS-COCO~\cite{coco} & SD2, SDXL, Firefly & \checkmark & 4.1k\\

SAGI- & MS-COCO~\cite{coco} & BrushNet
, PowerPaint, & & \\
\hfill SP~\cite{sagi25} & RAISE~\cite{raise15} & HD-Painter, ControlNet & \checkmark & 2k\\
\hfill FR~\cite{sagi25}& OpenImages~\cite{kuznetsova20} & Inpaint-Anything & $\times$ & 1.3k\\
SIDA~\cite{SIDA25} & MS-COCO~\cite{coco} & Latent Diffusion & $\times$ & 20k\\
MB ~\cite{MB24} & MS-COCO~\cite{coco} & Magic Brush & $\times$ & 465\\
FOSSIL- & MS-COCO~\cite{coco} & SD1.5, SD2, SD3, & \checkmark & \\
\hfill SP & & SDXL, Flux Kontex, & & 12.7k\\
\hfill FR & & Qwen Edit Image & $\times$ & \\
\hline
\end{tabular}
}
\end{table}

\subsubsection*{Sliding-window Inference}
\begin{table*}[t]
\caption{Average pixel-level IOU and F1 score comparison on image forgery localization. The best results are shown in bold, the second best are underlined.}
\centering
\resizebox{\textwidth}{!}{
\begin{tabular}{|r|cc|cc|cc|cc|cc|cc|cc|cc|cc|cc|}
\hline
& \multicolumn{2}{c|}{\textbf{Byd the Brush}} & \multicolumn{2}{c|}{\textbf{CocoGlide}} & \multicolumn{2}{c|}{\textbf{TGIF}} & \multicolumn{2}{c|}{\textbf{SAGI-FR}} & \multicolumn{2}{c|}{\textbf{SAGI-SP}} & \multicolumn{2}{c|}{\textbf{SIDA}} & \multicolumn{2}{c|}{\textbf{MB}} & \multicolumn{2}{c|}{\textbf{FOSSIL-SP}} & \multicolumn{2}{c|}{\textbf{FOSSIL-FR}} & \multicolumn{2}{c|}{\textbf{Average}}   \\
\multicolumn{1}{|c|}{\textbf{Model}} & \textbf{IOU} & \textbf{F1} & \textbf{IOU} & \textbf{F1} & \textbf{IOU} & \textbf{F1} & \textbf{IOU} & \textbf{F1} & \textbf{IOU} & \textbf{F1} & \textbf{IOU} & \textbf{F1} & \textbf{IOU} & \textbf{F1}  & \textbf{IOU} & \textbf{F1}  & \textbf{IOU} & \textbf{F1} & \textbf{IOU} & \textbf{F1} \\
\hline
DinoLizer (ours) & \textbf{0.38} & \textbf{0.50} & \textbf{0.74} & \textbf{0.82} & \underline{0.60} & \underline{0.68} & \textbf{0.29} & \textbf{0.40} & \textbf{0.56} & \textbf{0.64} & \textbf{0.60} & \textbf{0.69} & \textbf{0.65} & \textbf{0.76} & \textbf{0.66} & \textbf{0.72} & \textbf{0.45} & \textbf{0.53} & \textbf{0.55} & \textbf{0.64}\\
DINOv3-IML~\cite{dinov3} & 0.16 & 0.21 & \underline{0.48} & \underline{0.56} & 0.23 & 0.31 & \underline{0.23} & \underline{0.28} & 0.32 & 0.39 & \underline{0.23} & \underline{0.29} & 0.25 & 0.35 & 0.33 & 0.39 & \underline{0.16} & \underline{0.21} & 0.27 & 0.33\\
TruFor~\cite{turfor23} & \underline{0.22} & \underline{0.30} & 0.29 & 0.36 & \textbf{0.70} & \textbf{0.80} & 0.16 & 0.20 & \underline{0.48} & \underline{0.55} & 0.18 & 0.25 & \underline{0.28} & \underline{0.38} & \underline{0.63} & \underline{0.71} & 0.11 & 0.16 & \underline{0.34} & \underline{0.41}\\
SAFIRE~\cite{safire25} & 0.12 & 0.17 & 0.40 & 0.47 & 0.21 & 0.28 & 0.21 & \underline{0.28} & 0.27 & 0.33 & 0.21 & 0.28 & 0.21 & 0.29 & 0.29 & 0.36 & 0.15 & 0.21 & 0.24 & 0.30\\
ManTra-Net~\cite{mantra19} & 0.06 & 0.10 & 0.26 & 0.35 & 0.26 & 0.35 & 0.02 & 0.03 & 0.27 & 0.35 & 0.03 & 0.05 & 0.04 & 0.07 & 0.11 & 0.17 & 0.02 & 0.04 & 0.12 & 0.17\\
CAT-Net v2~\cite{catnetv2} & 0.13 & 0.20 & 0.17 & 0.24 & 0.14 & 0.23 & 0.03 & 0.05 & 0.11 & 0.18 & 0.08 & 0.13 & 0.02 & 0.04 & 0.07 & 0.12 & 0.01 & 0.02 & 0.09 & 0.14\\
SparseViT~\cite{sparse23} & 0.05 & 0.08 & 0.32 & 0.39 & 0.11 & 0.15 & 0.09 & 0.11 & 0.12 & 0.15 & 0.12 & 0.16 & 0.15 & 0.22 & 0.20 & 0.24 & 0.06 & 0.08 & 0.13 & 0.16\\ \hline
\end{tabular}
}
\label{tab:SOTAComparison}
\end{table*}

To localize inpainted regions in images of arbitrary resolution, we avoid down-sampling to preserve discriminative forensic cues. Instead, we apply a sliding window of size $w \times w = 504 \times 504$ with stride $s$ to the input image, producing a set of overlapping patches. For each patch, we compute logits via a linear classification head and sum them over all windows that cover a given pixel. Because the decision threshold remains fixed at $0$ (or $0.5$ in probability space), this summation is mathematically equivalent to averaging, as linear scaling does not alter the final classification. Grid search reveals that $s=128$ yields performance comparable to $s=64$ while doubling inference speed; thus, we adopt $s=128$ for all subsequent experiments. The window size $w$ matches the spatial resolution of our training set.

We observe that multiple overlapping windows improve prediction quality. Therefore, we upscale small test images to at least $1016 \times 1016$ pixels using bicubic interpolation to guarantee a $5 \times 5$ grid of windows. This ensures a diverse set of crops prior to averaging, further enhancing the accuracy of the localization map; the resulting masks are then resized to the original image dimensions.

While our detector is built upon the DINOv2-B \cite{bfree25} backbone, it departs fundamentally from its conventional use. Standard DINOv2-B averages probabilities from five global crops (four corners and the center) to produce a single binary decision. This strategy often fails when the inpainted region is small, as the global image statistics remain largely authentic. In contrast, our sliding-window approach yields a dense per-pixel prediction map that can reveal localized forgeries even when global statistics are unchanged.

\begin{figure*}[t]
    \centering
    \includegraphics[width=1.7\columnwidth]{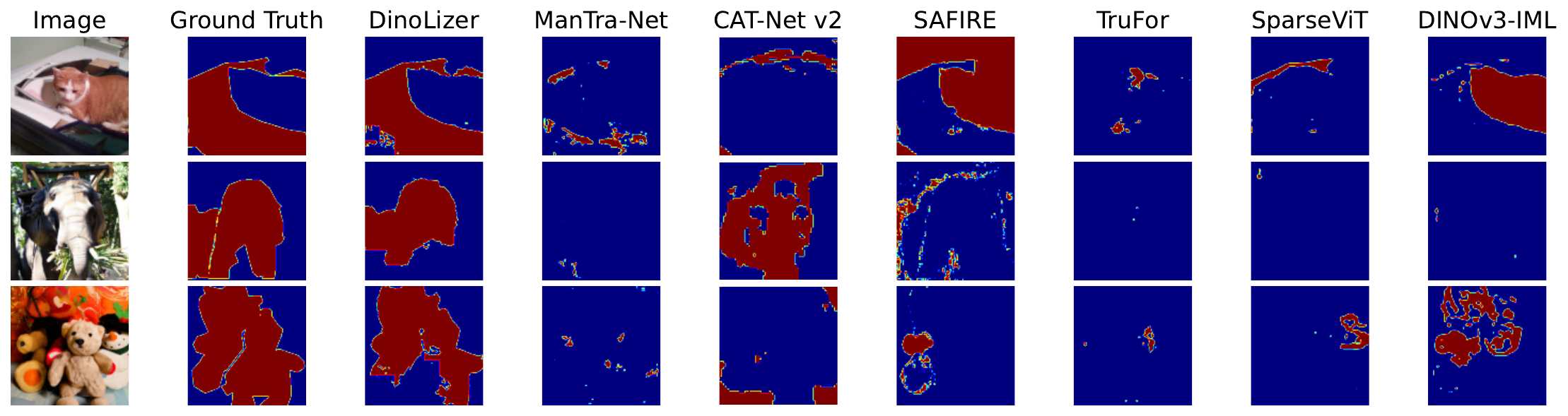}
    \caption{Comparison with existing methods on the CoCoGlide dataset.}
    \label{fig:comparison_CoCoGlide}
\end{figure*}

\section{Results}
\subsection{Experimental Setup}

\subsubsection*{\textbf{Training}}
We fine-tune the detector on the B-Free dataset, which contains 51k real and 309k synthetic images and a subset of the TGIF dataset
, which contains 118k synthetic images. The real samples are drawn from MS-COCO, whereas the synthetic images from the B-Free
dataset are reconstructed and inpainted variants of the real images produced with Stable Diffusion 2.1. The synthetic images from the TGIF dataset are generated using Stable Diffusion 2.1, XL, Flux 1 Dev, Flux 1 Schnell, and Flux 1 Fill.

DinoLizer is built upon a ViT-B/14 backbone with registers, refined on the B-Free dataset~\cite{bfree25}. We train the model with LORA adaptation and a linear classification head with 3 classes (real, VAE, and Inpainted) using the \texttt{AdamW} optimizer with a batch size of 64, an initial learning rate of $10^{-3}$, and a weight decay of $10^{-3}$. The learning rate is managed by a \texttt{lrdrop} scheduler, which halves the rate whenever the validation loss fails to improve for four consecutive epochs. Training is conducted for a maximum of 100 epochs on a single NVIDIA A6000 GPU (48GB), with early stopping triggered once the learning rate falls below $10^{-7}$. To enhance robustness against common image processing operations, we employ a diverse suite of augmentations via the Albumentations library on $504 \times 504$ images using Lanczos interpolation. These include random horizontal flips ($p=0.5$), 90$^\circ$ rotations ($p=1.0$), and a stochastic selection (each $p=0.25$) of \texttt{RandomResizedCrop}, \texttt{RandomCrop}, direct resizing, or upscaling by 200\%--300\% followed by \texttt{RandomCrop}. Furthermore, we incorporate uniform box blur ($p=0.1$), Gaussian noise ($\sigma \in [51, 112], p=0.1$), color jitter ($p=0.1$), and a two-stage JPEG compression process where the first stage uses a quality factor $QF \in \{40, \ldots, 100\}$ ($p=0.5$) and an optional second stage ($p=0.1$). During validation, we accelerate the evaluation process by applying only a center crop followed by an optional JPEG compression ($p=0.5$) with a randomly sampled quality factor.
\subsubsection*{\textbf{Testing}}
We evaluate our method on 5 public datasets of generative inpainting. In particular, we use the inpainted images from Beyond the Brush~\cite{btb24}, CocoGlide~\cite{turfor23}, TGIF~\cite{tgif24}, SAGI-SP~\cite{sagi25}, SAGI-FR~\cite{sagi25}, SIDA~\cite{SIDA25}, Magic Brush~\cite{MB24}, our benchmark FOSSIL-FR and FOSSIL-SP.
SP (splicing) and FR (fully regenerated) come from the same dataset, but we separate them in 2 situations. SP is the case where the region outside the mask is kept from the original image, while FR is the case where the region outside the mask is also regenerated by the inpainting model. Across
all datasets we consider roughly 48k images, see the details in Table~\ref{tab:datasets_overview}.

\subsubsection*{\textbf{Evaluation Metrics}}
We use IOU (Intersection over Union) and F1 score to evaluate the localization performance. A fixed threshold of 0.5 is used to binarize the predicted probability map.
We consider IOU equal to 1 when both the predicted and ground truth masks are empty (i.e., no inpainted pixels), which is necessary during training, as there are also fully pristine images and reconstructed images.

\subsection{State-of-the-art Comparison}
\subsubsection*{\textbf{Qualitative Results}}

\begin{figure*}[t]
    \centering
    \includegraphics[width=1.8\columnwidth]{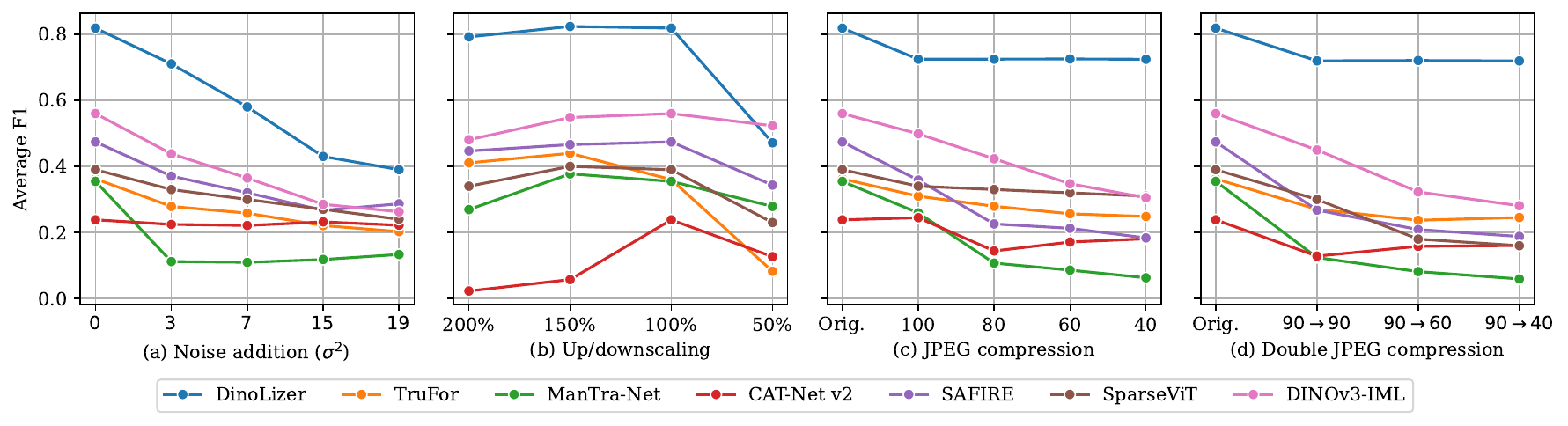}
    \caption{F1 score performance under different types of perturbations on the CocoGlide dataset: (a) gaussian noise, (b) resizing, (c) JPEG compression, and (d) double JPEG compression.}

    \label{fig:robustness}
\end{figure*}

We evaluate DinoLizer against a representative set of state-of-the-art inpainting-forgery detectors, employing only the publicly released weights. The competing methods are TruFor~\cite{turfor23}, SAFIRE~\cite{safire25}, ManTra-Net~\cite{mantra19}, CAT-Net v2~\cite{catnetv2}, and SparseViT~\cite{sparse23}. All models are used off-the-shelf without any additional re-training.

Table~\ref{tab:SOTAComparison} summarizes the quantitative results. DinoLizer attains the highest F1 score across all datasets, with the sole exception of the TGIF benchmark where TruFor slightly surpasses it. On average, our approach outperforms the other detectors by at least 20\% in F1.
We can also notice that although TruFor performs well on FOSSIL-SP, its accuracy drops significantly on FOSSIL-FR (with regenerated background), while our method is more robust to both situations.

Figure \ref{fig:comparison_CoCoGlide} presents a qualitative comparison of DinoLizer against the SOTA on the CoCoGlide dataset, demonstrating that our method can localize well the inpainted regions.

\subsubsection*{\textbf{Robustness Evaluation}}

We evaluate the robustness of the detectors against four common post-processing operations: \textbf{JPEG compression} with quality factors $QF \in \{100, 80, 60, 40\}$; \textbf{double JPEG compression} (first pass $QF_1=90$, second pass $QF_2 \in \{90, 60, 40\}$); \textbf{bicubic resizing} to 50\%, 150\%, and 200\% of the original resolution; and \textbf{additive Gaussian noise} with variance $\sigma^2 \in \{3, 7, 15, 19\}$.

Figure~\ref{fig:robustness} illustrates the robustness of DinoLizer across various perturbation types. Across nearly all categories, DinoLizer significantly outperforms the baseline methods. Notably, our model exhibits remarkable stability under JPEG (double) compression (Figure~\ref{fig:robustness}c, d), maintaining a consistent F1 score from QF=100 to QF=40. While DinoLizer shows sensitivity to Gaussian noise (Figure~\ref{fig:robustness}a), it remains the top-performing model across all noise levels; however, due to the notable drops in performance, we identify this sensitivity as an important area for further investigation in future work. Furthermore, in the up/downsampling task (Figure~\ref{fig:robustness}b), DinoLizer maintains a substantial lead over most baselines, with the sole exception of the 50\% downsampling case, where it slightly underperforms compared to DINOv3-IML.

\section{Ablation Studies}
\subsection{LORA Adaptation}
\begin{table}[t]
\caption{Ablation study on the impact of LoRA adaptation, pretrain, and multiclass on the model's performance}
\label{tab:ablation_lora}
\centerline{
\begin{tabular}{|l|l|c|c|c|}
\hline
\textbf{Pretrain} & \textbf{Setup} & \textbf{\# Classes} & \textbf{IOU} & \textbf{F1} \\ \hline
& 1 linear layer & 2 & 0.31 & 0.41 \\
& 2 transformer blocks & 2  & 0.40 & 0.50 \\
BFree \cite{bfree25} & & 3 & 0.49 & 0.58 \\
& LoRA Adaptation & 2 & 0.51 & 0.60 \\
& & 3 & \textbf{0.55} & \textbf{0.64} \\
\hline
LVD-142M \cite{dinov2} & 2 transformer blocks & 3 & 0.44 & 0.54 \\
& LoRA Adaptation & 3 & 0.54 & 0.62 \\
\hline
\end{tabular}
}
\end{table}
\vspace{-1mm}

We evaluate the effectiveness of LoRA in our model by comparing its performance with a frozen model where we train with a 2-transformer-block head on top of the frozen model. The results, in Table~\ref{tab:ablation_lora}, indicate that LoRA adaptation significantly enhances the model's ability to learn discriminative features for inpainting localization, leading to improved performance across all evaluation metrics while reducing the number of trainable parameters (6\% in IOU for the 3-class setting and 11\% in IOU for the 2-class setting).

\subsection{3 classes vs 2 classes}
We conduct an ablation study to evaluate the impact of our proposed 3-class training strategy (Real, VAE, and Inpainted) compared to the conventional 2-class setup (Real and Inpainted). Specifically, we investigate two primary configurations: 2-class and 3-class settings, both of which utilize LoRA adaptation or two additional transformer blocks on top of the frozen backbone. Furthermore, we establish a baseline by evaluating a model with a frozen backbone and a simple linear head for the 2-class classification task. For the 2-class setup, we use the Weighted Binary Cross-Entropy loss. The three-class setup follows that proposed in Section~\ref{sec:3class}.

Our results in Table~\ref{tab:ablation_lora} demonstrate that the 3-class configuration significantly enhances the model's ability to differentiate between real/VAE and inpainted regions, leading to superior localization performance across all evaluation metrics.

\subsection{Pretrained vs Not pretrained}


We investigate the impact of fine-tuning DINOv2 on the B-Free dataset by comparing our model's performance with and without this step. As shown in Table~\ref{tab:ablation_lora}, while this fine-tuning yields almost negligible improvements in overall performance for our primary configuration (LORA), it provides a more noticeable gain when using only two transformer blocks.

\section*{Acknowledgment}
This work was also supported by a French government grant managed by the {\it Agence Nationale de la Recherche} under the France 2030 program, reference ANR-22-PECY0011 and the project ANR-23-IAS4-0004.

\section{Conclusion}
We propose a novel framework that leverages the versatile DINO family of Vision Transformers to localize generated image regions, achieving state-of-the-art performance across multiple benchmarks.
Additionally, we introduce a new dataset, \textbf{FOSSIL}, which incorporates contemporary generative inpainting methods, such as Flux Kontex and Qwen Edit Image, providing a more challenging evaluation scenario for existing detectors.

Our methodology addresses the localization challenge in inpainting forgeries by explicitly modeling auto-encoded regions as a distinct class alongside real and inpainted areas. By employing Low-Rank Adaptation, we are able to fine-tune the model with only 2.4M trainable parameters while maintaining state-of-the-art performance.
The proposed solution is highly computationally efficient: by leveraging DINO's patch tokens and a sliding window approach, both training and inference can be performed on a single GPU, even for large images.
Furthermore, our comprehensive data augmentation strategy ensures robustness against common image post-processing operations, most notably JPEG (double) compression. This resilience is particularly valuable given the diverse processing pipelines that forgeries often undergo before analysis.

\bibliographystyle{IEEEtran}
\bibliography{IEEEabrv}

\begin{thebibliography}{10}
\providecommand{\url}[1]{#1}
\csname url@samestyle\endcsname
\providecommand{\newblock}{\relax}
\providecommand{\bibinfo}[2]{#2}
\providecommand{\BIBentrySTDinterwordspacing}{\spaceskip=0pt\relax}
\providecommand{\BIBentryALTinterwordstretchfactor}{4}
\providecommand{\BIBentryALTinterwordspacing}{\spaceskip=\fontdimen2\font plus
\BIBentryALTinterwordstretchfactor\fontdimen3\font minus \fontdimen4\font\relax}
\providecommand{\BIBforeignlanguage}[2]{{%
\expandafter\ifx\csname l@#1\endcsname\relax
\typeout{** WARNING: IEEEtran.bst: No hyphenation pattern has been}%
\typeout{** loaded for the language `#1'. Using the pattern for}%
\typeout{** the default language instead.}%
\else
\language=\csname l@#1\endcsname
\fi
#2}}
\providecommand{\BIBdecl}{\relax}
\BIBdecl

\bibitem{rombach2022high}
R.~Rombach, A.~Blattmann, D.~Lorenz, P.~Esser, and B.~Ommer, ``High-resolution image synthesis with latent diffusion models,'' in \emph{Proceedings of the IEEE/CVF conference on computer vision and pattern recognition (CVPR)}, 2022, pp. 10\,684--10\,695.

\bibitem{nichol2021glide}
A.~Nichol, P.~Dhariwal, A.~Ramesh, P.~Shyam, P.~Mishkin, B.~McGrew, I.~Sutskever, and M.~Chen, ``Glide: Towards photorealistic image generation and editing with text-guided diffusion models,'' \emph{arXiv preprint arXiv:2112.10741}, 2021.

\bibitem{suvorov2022resolution}
R.~Suvorov, E.~Logacheva, A.~Mashikhin, A.~Remizova, A.~Ashukha, A.~Silvestrov, N.~Kong, H.~Goka, K.~Park, and V.~Lempitsky, ``Resolution-robust large mask inpainting with fourier convolutions,'' in \emph{Proceedings of the IEEE/CVF winter conference on applications of computer vision}, 2022, pp. 2149--2159.

\bibitem{ju2024brushnet}
X.~Ju, X.~Liu, X.~Wang, Y.~Bian, Y.~Shan, and Q.~Xu, ``Brushnet: A plug-and-play image inpainting model with decomposed dual-branch diffusion,'' in \emph{European Conference on Computer Vision}.\hskip 1em plus 0.5em minus 0.4em\relax Springer, 2024, pp. 150--168.

\bibitem{zhuang2024task}
J.~Zhuang, Y.~Zeng, W.~Liu, C.~Yuan, and K.~Chen, ``A task is worth one word: Learning with task prompts for high-quality versatile image inpainting,'' in \emph{European Conference on Computer Vision}.\hskip 1em plus 0.5em minus 0.4em\relax Springer, 2024, pp. 195--211.

\bibitem{manukyan2023hd}
H.~Manukyan, A.~Sargsyan, B.~Atanyan, Z.~Wang, S.~Navasardyan, and H.~Shi, ``Hd-painter: high-resolution and prompt-faithful text-guided image inpainting with diffusion models,'' in \emph{The Thirteenth International Conference on Learning Representations (ICLR)}, 2023.

\bibitem{xie2023smartbrush}
S.~Xie, Z.~Zhang, Z.~Lin, T.~Hinz, and K.~Zhang, ``Smartbrush: Text and shape guided object inpainting with diffusion model,'' in \emph{Proceedings of the IEEE/CVF conference on computer vision and pattern recognition}, 2023, pp. 22\,428--22\,437.

\bibitem{zhang2023adding}
L.~Zhang, A.~Rao, and M.~Agrawala, ``Adding conditional control to text-to-image diffusion models,'' in \emph{Proceedings of the IEEE/CVF international conference on computer vision (ICCV)}, 2023, pp. 3836--3847.

\bibitem{turfor23}
F.~Guillaro, D.~Cozzolino, A.~Sud, N.~Dufour, and L.~Verdoliva, ``Trufor: Leveraging all-round clues for trustworthy image forgery detection and localization,'' in \emph{Proceedings of the IEEE/CVF conference on computer vision and pattern recognition (CVPR)}, 2023, pp. 20\,606--20\,615.

\bibitem{safire25}
M.-J. Kwon, W.~Lee, S.-H. Nam, M.~Son, and C.~Kim, ``Safire: Segment any forged image region,'' vol.~39, no.~4, 2025, pp. 4437--4445.

\bibitem{mantra19}
W.~A. Yue~Wu and P.~Natarajan, ``Mantra-net: Manipulation tracing network for detection and localization of image forgeries with anomalous features,'' in \emph{The IEEE Conference on Computer Vision and Pattern Recognition (CVPR)}, 2019.

\bibitem{catnetv2}
M.-J. Kwon, S.-H. Nam, I.-J. Yu, H.-K. Lee, and C.~Kim, ``Learning jpeg compression artifacts for image manipulation detection and localization,'' \emph{International Journal of Computer Vision}, vol. 130, no.~8, pp. 1875--1895, 2022.

\bibitem{su25}
L.~Su, X.~Ma, X.~Zhu, C.~Niu, Z.~Lei, and J.-Z. Zhou, ``Can we get rid of handcrafted feature extractors? sparsevit: Nonsemantics-centered, parameter-efficient image manipulation localization through spare-coding transformer,'' in \emph{Proceedings of the AAAI Conference on Artificial Intelligence}, vol.~39, no.~7, 2025, pp. 7024--7032.

\bibitem{sparse23}
X.~Chen, Z.~Liu, H.~Tang, L.~Yi, H.~Zhao, and S.~Han, ``Sparsevit: Revisiting activation sparsity for efficient high-resolution vision transformer,'' in \emph{Proceedings of the IEEE/CVF Conference on Computer Vision and Pattern Recognition}, 2023, pp. 2061--2070.

\bibitem{hu2021loralowrankadaptationlarge}
\BIBentryALTinterwordspacing
E.~J. Hu, Y.~Shen, P.~Wallis, Z.~Allen-Zhu, Y.~Li, S.~Wang, L.~Wang, and W.~Chen, ``Lora: Low-rank adaptation of large language models,'' 2021. [Online]. Available: \url{https://arxiv.org/abs/2106.09685}
\BIBentrySTDinterwordspacing

\bibitem{dinov2}
M.~Oquab, T.~Darcet, T.~Moutakanni, H.~Vo, M.~Szafraniec, V.~Khalidov, P.~Fernandez, D.~Haziza, F.~Massa, A.~El-Nouby \emph{et~al.}, ``Dinov2: Learning robust visual features without supervision,'' \emph{Transactions on Machine Learning Research Journal}, pp. 1--31, 2024.

\bibitem{bfree25}
F.~Guillaro, G.~Zingarini, B.~Usman, A.~Sud, D.~Cozzolino, and L.~Verdoliva, ``A bias-free training paradigm for more general ai-generated image detection,'' in \emph{Proceedings of the Computer Vision and Pattern Recognition Conference (CVPR)}, 2025, pp. 18\,685--18\,694.

\bibitem{strudel21}
R.~Strudel, R.~Garcia, I.~Laptev, and C.~Schmid, ``Segmenter: Transformer for semantic segmentation,'' in \emph{Proceedings of the IEEE/CVF international conference on computer vision (ICCV)}, 2021, pp. 7262--7272.

\bibitem{dinov3}
\BIBentryALTinterwordspacing
J.~Yu, Q.~Feng, Z.~Wang, and X.~Ma, ``Dinov3 beats specialized detectors: A simple foundation model baseline for image forensics,'' 2026. [Online]. Available: \url{https://arxiv.org/abs/2604.16083}
\BIBentrySTDinterwordspacing

\bibitem{coco}
\BIBentryALTinterwordspacing
T.-Y. Lin, M.~Maire, S.~Belongie, L.~Bourdev, R.~Girshick, J.~Hays, P.~Perona, D.~Ramanan, C.~L. Zitnick, and P.~Dollár, ``Microsoft coco: Common objects in context,'' 2015. [Online]. Available: \url{https://arxiv.org/abs/1405.0312}
\BIBentrySTDinterwordspacing

\bibitem{corvi2025seeing}
R.~Corvi, D.~Cozzolino, E.~Prashnani, S.~De~Mello, K.~Nagano, and L.~Verdoliva, ``Seeing what matters: Generalizable ai-generated video detection with forensic-oriented augmentation,'' \emph{arXiv preprint arXiv:2506.16802}, 2025.

\bibitem{chen2025dual}
R.~Chen, J.~Xi, Z.~Yan, K.-Y. Zhang, S.~Wu, J.~Xie, X.~Chen, L.~Xu, I.~Guan, T.~Yao \emph{et~al.}, ``Dual data alignment makes ai-generated image detector easier generalizable,'' \emph{arXiv preprint arXiv:2505.14359}, 2025.

\bibitem{btb24}
G.~Bertazzini, C.~Albisani, D.~Baracchi, D.~Shullani, and A.~Piva, ``Beyond the brush: Fully-automated crafting of realistic inpainted images,'' in \emph{2024 IEEE International Workshop on Information Forensics and Security (WIFS)}, 2024, pp. 1--6.

\bibitem{fooocus23}
L.~Zhang, ``Fooocus,'' \url{https://github.com/lllyasviel/Fooocus}, 2023, online; accessed 2025-11-11.

\bibitem{glide22}
\BIBentryALTinterwordspacing
A.~Q. Nichol, P.~Dhariwal, A.~Ramesh, P.~Shyam, P.~Mishkin, B.~Mcgrew, I.~Sutskever, and M.~Chen, ``{GLIDE}: Towards photorealistic image generation and editing with text-guided diffusion models,'' in \emph{Proceedings of the 39th International Conference on Machine Learning}, ser. Proceedings of Machine Learning Research, K.~Chaudhuri, S.~Jegelka, L.~Song, C.~Szepesvari, G.~Niu, and S.~Sabato, Eds., vol. 162.\hskip 1em plus 0.5em minus 0.4em\relax PMLR, 17--23 Jul 2022, pp. 16\,784--16\,804. [Online]. Available: \url{https://proceedings.mlr.press/v162/nichol22a.html}
\BIBentrySTDinterwordspacing

\bibitem{tgif24}
H.~Mareen, D.~Karageorgiou, G.~V. Wallendael, P.~Lambert, and S.~Papadopoulos, ``Tgif: Text-guided inpainting forgery dataset,'' in \emph{2024 IEEE International Workshop on Information Forensics and Security (WIFS)}, 2024, pp. 1--6.

\bibitem{sagi25}
P.~Giakoumoglou, D.~Karageorgiou, S.~Papadopoulos, and P.~C. Petrantonakis, ``Sagi: Semantically aligned and uncertainty guided ai image inpainting,'' in \emph{International Conference on Computer Vision (ICCV)}, 2025.

\bibitem{raise15}
\BIBentryALTinterwordspacing
D.-T. Dang-Nguyen, C.~Pasquini, V.~Conotter, and G.~Boato, ``Raise: a raw images dataset for digital image forensics,'' in \emph{Proceedings of the 6th ACM Multimedia Systems Conference}, ser. MMSys '15.\hskip 1em plus 0.5em minus 0.4em\relax New York, NY, USA: Association for Computing Machinery, 2015, p. 219–224. [Online]. Available: \url{https://doi.org/10.1145/2713168.2713194}
\BIBentrySTDinterwordspacing

\bibitem{kuznetsova20}
A.~Kuznetsova, H.~Rom, N.~Alldrin, J.~Uijlings, I.~Krasin, J.~Pont-Tuset, S.~Kamali, S.~Popov, M.~Malloci, A.~Kolesnikov \emph{et~al.}, ``The open images dataset v4: Unified image classification, object detection, and visual relationship detection at scale,'' \emph{International journal of computer vision}, vol. 128, no.~7, pp. 1956--1981, 2020.

\bibitem{SIDA25}
\BIBentryALTinterwordspacing
Z.~Huang, J.~Hu, X.~Li, Y.~He, X.~Zhao, B.~Peng, B.~Wu, X.~Huang, and G.~Cheng, ``Sida: Social media image deepfake detection, localization and explanation with large multimodal model,'' 2025. [Online]. Available: \url{https://arxiv.org/abs/2412.04292}
\BIBentrySTDinterwordspacing

\bibitem{MB24}
\BIBentryALTinterwordspacing
K.~Zhang, L.~Mo, W.~Chen, H.~Sun, and Y.~Su, ``Magicbrush: A manually annotated dataset for instruction-guided image editing,'' 2024. [Online]. Available: \url{https://arxiv.org/abs/2306.10012}
\BIBentrySTDinterwordspacing

\end{thebibliography}

\end{document}